\documentclass[letterpaper]{article} 
\usepackage{aaai24}  
\usepackage{times}  
\usepackage{helvet}  
\usepackage{courier}  
\usepackage[hyphens]{url}  
\usepackage{graphicx} 
\usepackage{natbib}  
\usepackage{caption} 
\usepackage{algorithm}
\usepackage{algorithmic}
\usepackage{subcaption}
\usepackage{amsmath}
\usepackage{amsthm}
\usepackage{booktabs}
\usepackage{multirow}
\usepackage{diagbox}
\usepackage{makecell}
\nocopyright

\usepackage{newfloat}
\usepackage{listings}
\DeclareCaptionStyle{ruled}{labelfont=normalfont,labelsep=colon,strut=off} 
\floatstyle{ruled}
\newfloat{listing}{tb}{lst}{}
\floatname{listing}{Listing}
\title{Distinctive Self-Similar Object Detection}
\author{
    Zeyu Shangguan,
    Bocheng Hu\equalcontrib,
    Guohua Dai\equalcontrib,
    Yuyu Liu,
    Darun Tang,
    Xingqun Jiang
}
\affiliations{
    BOE Technology Group Co., Ltd.\\

    No.09 Dize Road, Daxing District,
    Beijing, China, 100176\\
    \{shangguanzeyu, hubocheng, daiguohua, liuyuyu, tangdarun, jiangxingqun\}@boe.com.cn
}

\usepackage{bibentry}

\begin{document}

\maketitle

\begin{abstract}
Deep learning-based object detection has demonstrated a significant presence in the practical applications of artificial intelligence. However, objects such as fire and smoke, pose challenges to object detection because of their non-solid and various shapes, and consequently difficult to truly meet requirements in practical fire prevention and control. In this paper, we propose that the distinctive fractal feature of self-similar in fire and smoke can relieve us from struggling with their various shapes. To our best knowledge, we are the first to discuss this problem. In order to evaluate the self-similarity of the fire and smoke and improve the precision of object detection, we design a semi-supervised method that use Hausdorff distance to describe the resemblance between instances. Besides, based on the concept of self-similar, we have devised a novel methodology for evaluating this particular task in a more equitable manner. We have meticulously designed our network architecture based on well-established and representative baseline networks such as YOLO and Faster R-CNN. Our experiments have been conducted on publicly available fire and smoke detection datasets, which we have thoroughly verified to ensure the validity of our approach. As a result, we have observed significant improvements in the detection accuracy.
\end{abstract}

\section{Introduction}
\label{sec: Introduction}

With the development of science and technology, object detection based on artificial intelligence has gradually replaced traditional object detection because of its excellent performance, and has made excellent achievements in all walks of life~\cite{KAUR2023103812}. As an important application of object detection based on artificial intelligence, fire detection has great utility and practical demand for preventing conflagration and other scenarios. The United Nations Environment Programme released a report titled "Spreading like Wildfire–The Rising Threat of Extraordinary Landscape Fires" in early 2022~\cite{UNEP_report}. It states that climate change and land-use changes are expected to lead to more frequent and intense wildfires. On a global scale, extreme fire incidents will increase by 14\% by 2030, 30\% by the end of 2050 and 50\% by the end of this century. For example, the wildfire that occurred on July 30, 2022, named the Kinney Fire, is considered to be the largest one in California this year. According to data from the National Interagency Coordination Center, the area of land burned by wildfires in California has exceeded 12,500 square kilometers this year, equivalent to the size of 10 New York City areas. The report calls for a fundamental shift in how governments invest in addressing wildfires, from spending on "late response and adaptation" to investing in "prevention and preparedness" at the front end. Fire and smoke not only causes serious economic losses, but also poses a serious threat to personal safety. The initial purpose of fire and smoke detection is to predict the occurrence of fire or smoke in advance to achieve early warning effect~\cite{PAYRA2023239}.

\begin{figure}[tbp]
	\centering
	\includegraphics[width=85mm]{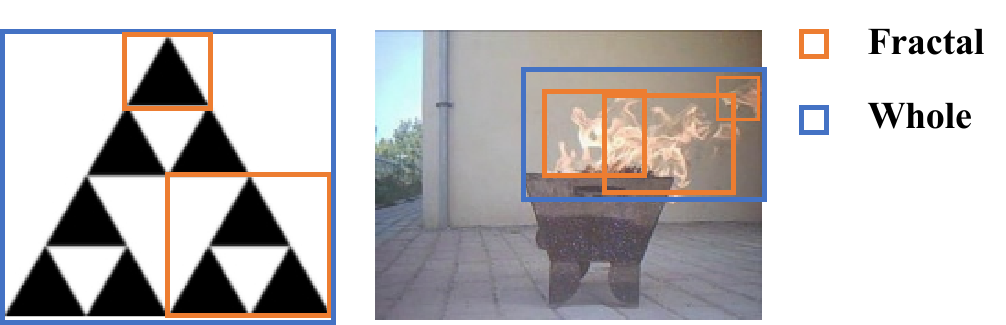}
	\caption{The left part is a Sierpinski triangle, which is a standard self-similar pattern. While the right part demonstrates the same patterns on the fire. The self-similar feature of fire and smoke has provided us with a promising breakthrough point to enhance the recognition ability.}
	\label{fig:intro}
\end{figure}

However, compared with other daily necessities, it is more difficult to detect the fire and smoke objects because of their uncertain shape and semitransparent appearance, which is challenging in achieving a consolidated standard for both data labeling and model evaluation processes for fire and smoke detection tasks. While using existing artificial intelligence-based object detection for fire and smoke detection, when only a part of the fire or smoke is detected, the model is likely to predict it as background or false positive sample. However, in reality, a part of the fire or smoke is equal to the whole fire or smoke and therefore should be considered foreground or true positive sample. This obviously goes against the original intention of the network design, which leads to the confusion in the process of data set preparation: there will be non-unique criteria for data labeling, and the classic model evaluation could not properly present the potential true positive samples. 

Considering all the above factors and the shortcomings of existing object detection models, There are two major challenges: (1) The model has the ability to recognize that a part of the fire or smoke is equal to a whole fire or smoke and make accurate predication. (2) To establish a model evaluation mechanism suitable for such tasks.

Through the observation of a large number of flame and smoke images, we found that both fire and smoke have self-similarity with any part of it, as shown in Figure~\ref{fig:intro}: natural objects such as fire and smoke generally have self-similar geometric characteristics, such as the standard Sierpinski triangle, and any fractal is highly geometrically similar to the whole. In other words, the basic truth is that any part of the fire and smoke can also be seen as fire or smoke. Thus, during the model training, if the predicted part has the fractal similarity features with the whole target, it should be considered that the model makes a correct detection.


To verify the rationality of the above hypothesis, we implemented an object detection algorithm based on self-similarity metrics, using YOLOv5 and Faster R-CNN network structures. During feature extraction stages, We propose to use the Hausdorff distance to measure the self-similarity of fire fractals and fire as a whole, and then design loss formulas based on this. As far as we know, this is the first time that self-similar objects have been explored in this context.

In addition, we designed a test evaluation index based on self-similarity characteristics that is different from the standard AP50, and verified it on two publicly available datasets of fireworks detection. Finally, our method is proven to be effective and superior.

Our main contributes and novelties include:
\begin{itemize}
	\item We propose to utilize the Hausdorff distance as a way to evaluate the self-similar feature and apply it in the training process.
	\item We build a novel paradigm for evaluating the precision of the self-similar objects.
	\item We achieve significant improvement on common used data sets.
\end{itemize}

\section{Related Works}
\label{sec: Related Works}

\subsection{Object Detection}
\label{sec: Object Detection}

Most object detection tasks usually are separated as one-stage and two-stage method~\cite{li-et-al:deep}. The one-stage methods, such as YOLO~\cite{7780460,bochkovskiy-et-al:YOLOv4}, SSD~\cite{Liu_2016}, RetinaNET~\cite{lin2018focal}, EfficientDet~\cite{tan2020efficientdet}, YOLO-Ret~\cite{ganesh-et-al:YOLO-ReT}, and YOLOX~\cite{ge-et-al:YOLOX}, are getting the bounding box and classify directly without the region proposal. The two-stage methods, such as R-CNN~\cite{girshick2014rich},  Fast RCNN~\cite{girshick2015fast}, Faster RCNN~\cite{ren2016faster}, Mask RCNN~\cite{he2018mask}, are to firstly get the region proposal, and then do the classification and the bounding box regression. Each of these methods has its strengths and weaknesses, and the choice of which one to use depends on the specific requirements of the application. The one-stage methods are generally favored for detection tasks that require real-time or nearly real-time processing, while the two-stage methods deliver better accuracy at the cost of increased computational complexity. Therefore, in this paper we pick both the representative one-stage (i.e. YOLO) and two-stage methods (i.e. Faster R-CNN) to verify the generalization of our method.

\subsection{Fire and Smoke Detection}
\label{sec: Fire and Smoke Detection}


\begin{figure}[tbp]
	\centering
	\includegraphics[width=50mm]{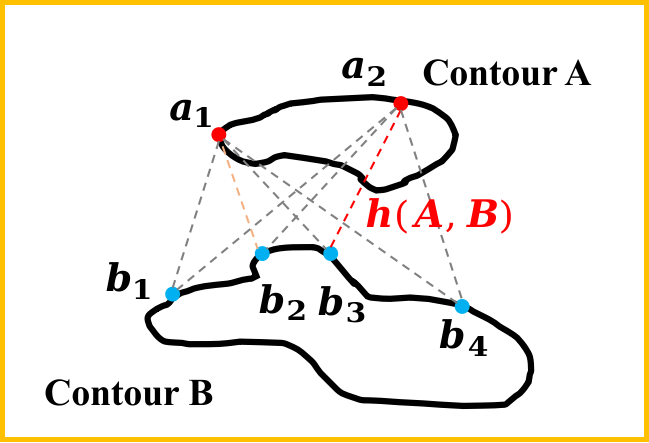}
	\caption{$h(A,B)$ is the one way Hausdorff distance from contour A to contour B.}
	\label{fig:hausdorff}
\end{figure}

\begin{figure*}[tb]
	\centering
    \begin{subfigure}{\linewidth}
		\centering
	    \includegraphics[width=140mm]{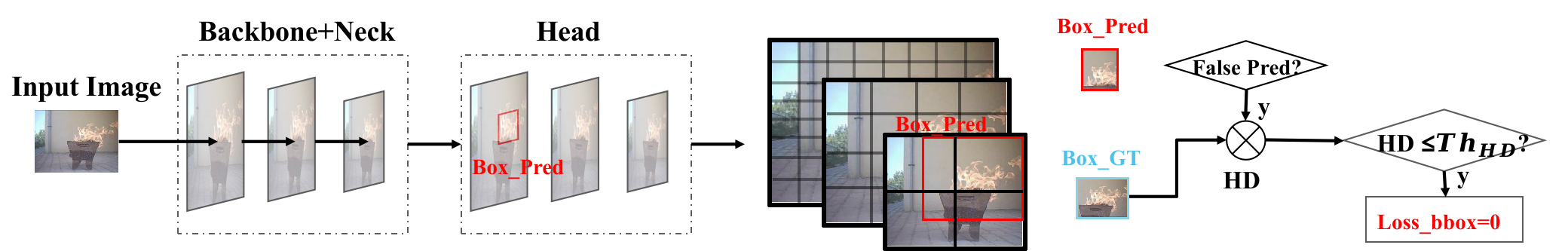}
		\caption{Our improved YOLO network.}
		\label{fig:yolo}
	\end{subfigure}
 
    \begin{subfigure}{\linewidth}
		\centering
	\includegraphics[width=175mm]{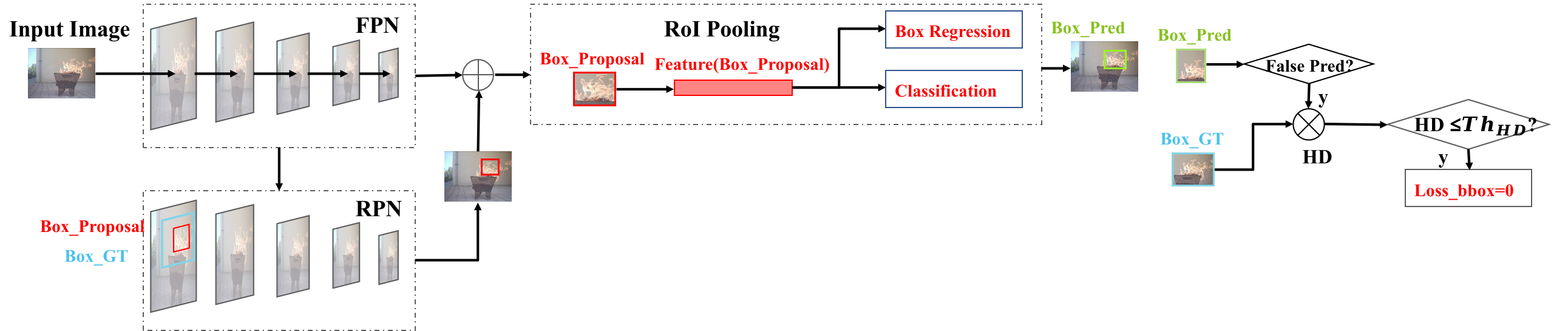}
	\caption{Our improved Faster R-CNN network.}
	\label{fig:frcn}
	\end{subfigure}
	\caption{The self-similar measure module is added after the detection head for YOLO and RoI pooling for Faster R-CNN. Once the predicted area is similar to the ground truth in terms of Hausdorff distance, we set the corresponding box regression loss as zero.}
	\label{fig:pipeline}
\end{figure*}

Current fire recognition methods can be classified as traditional vision-based method, CNN-based classification, R-CNN based object detection, video-based analysis~\cite{kim-et-al:a,jeong-et-al:light,xu-et-al:advances} and instance segmentation~\cite{dunnings-et-al:experimentally}. The traditional vision-based method is to analyze the pixel color. ~\cite{chen-et-al:digital} has found that the fire usually has a unique frequency of flick, therefore, an area can be inferred as fire if it has a regular mode of pixel color change. ~\cite{muhammad-et-al:convolutional} tries to use CNN-based classification to recognize whether an image contains fire.  ~\cite{li-et-al:image} uses object detection method to get the bounding boxes of the fire and smoke area. This method may consume human work to label the ground truth bounding boxes for each image, but can get the accurate position of the fire and smoke, which is beneficial for some commercial applications such as vehicle fire alarm, forest fire alarm, kitchen fire and smoke alarm, etc. ~\cite{su14094930} proposes an improved Fire-YOLO deep learning algorithm, which expands the feature extraction network from three dimensions: depth, width, and resolution, enhances feature propagation of fire small targets identification, improves network performance, and reduces model parameters. ~\cite{9356284} proposes reduced complexity compact CNN architectures (NasNet-A-OnFire and ShuffleNetV2-OnFire), improves upon the current state-of-the-art solution for fire detection, and optimizes the computational efficiency to 95\% for full-frame binary classification and 97\% for superpixel localization. Related works also improve upon the current state-of-the-art solution for fire detection using CNN-based methods, such as ~\cite{li-et-al:image,liu2021real}. ~\cite{yan2023transmissionguided} introduces a Bayesian generative model to solve the problems of limited training data and ground-truth labeling difficulty, and simultaneously estimate the posterior distribution of model parameters and its predictions. To solve the lack of training data, ~\cite{ZHANG2018441} produces synthetic smoke images to train the Faster R-CNN network by inserting real smoke or stimulative smoke into forest background to achieve data augmentation. We concentrate on object detection-based method because it could properly balancing accurate localization and real-time inferencing.

\subsection{Self-similar}
\label{sec: Self-similar}

The self-similarity refers to a feature where the parts of objects exhibit similar patterns or characteristics of the whole complex structure. It is a widely seen phenomenon in nature: things such as snowflake, tree branch and coastline are all self-similar since each part of them are visually similar to themselves. In the mathematics field of fractal geometry, self-similarity could be observed in fractals, which are complex geometric shapes or patterns that display the same structure when viewed at different scales.  The Hausdorff distance ($HD$) is widely used to describe the shape similarity of objects in AI Engineering field~\cite{silva-et-al:fractal, rapaport-et-al:on}, and therefore the most intuitive way to metric the similarity between fractal and whole, namely self-similarity. 




\begin{equation}
\begin{split}
HD(A,B) = max\left\lbrace h(A,B),h(B,A)\right\rbrace  \\
h(A,B) = max\left\lbrace min\left\lbrace \lVert a-b\rVert \right\rbrace \right\rbrace
\end{split}
\label{eq:hausdorff distance}
\end{equation}

\begin{itemize}
	\item $HD(A,B)$: the Hausdorff distance between set A to B.
	\item $h(A,B)$: the Hausdorff distance from set A to B.
\end{itemize}

Figure~\ref{fig:hausdorff} demonstrates the calculation of Hausdorff distance when evaluating the similarity between two images. It works by calculating the maximum value of the minimum distances of the points from one contour to another contour. The definition is as Equation~\ref{eq:hausdorff distance}. The lower the value of $HD$, the closer the two sets of contours will be in terms of their similarity.

\subsection{Datasets}
\label{sec: Datasets}

In our paper, we utilize widely accepted and commonly used datasets for fire and smoke detection, such as DFS~\cite{wu2022dataset} and FireNet~\cite{Moses19}, for evaluation purposes. DFS includes 9462 images in 4 categories: fire, smoke, other (similar to fire) and background, but it does not provide the sample partition. Additionaly, it classifies the images as large and middle according the size of objects. While FireNet contains 412 images for training and 90 images for validation, and it only have one category of fire. In order to evaluate fairly, we finish the partition seriously on DFS dataset and open source the partition method, details will be described in Section~\ref{sec: Experiments}.

\section{Proposed Methods}
\label{sec:Proposed Methods}

In order to establish the universality of our proposal, we have implemented it on two widely-used object detection networks: YOLOv5 and Faster R-CNN. Details will be presented in this section.

\subsection{Self-similar Loss Function}
\label{sec: Self-similar Loss Function}

The core idea is, for the self-similar objects, if the predicted areas contain any fractions that geometrically similar to the whole area, then these predicted results refer to correct predictions. In this situation, even though the bounding box of the predicted fraction has low intersection over union (IoU) with respect to the ground truth, we still treat it as true positive sample, and consequently assign it with low box regression loss. This is a semi-supervised method as we are committed to find potential positive fractals to enhance the recognition ability of the model. 

Specifically, for YOLO network, as shown in Figure ~\ref{fig:yolo}, the training image is firstly sent into the backbone-neck-head module to extract features in different scales, and then box predictions are performed upon the feature maps to generate predict boxes and predict their categories. In order to distinguish the potential fractions ($Frac_p$) from the predicted true negative boxes, we firstly obtain all true negative boxes by filtering out the predicted boxes that have high IoU. Then we crop the predicted area and the corresponding ground truth area from the origin image so that we metric the similarity of these two images through Hausdorff distance (HD). For those predicted areas that have small $HD$ with respect to ground truth area that lower than a fixed Hausdorff distance threshold ($Th_{HD}$), we consider them as a $Frac_p$. Accordingly, the box regression loss is hard clipped to zero. The corresponding loss function is described in Equation~\ref{eq:yolo loss function}.

\begin{equation}
L_{bbox}\!=\!I\{HD\!\leq\!Th_{HD}\} \cdot \{\!1\!-\!IOU\!+\!\frac{\rho^{2}(b,b^{gt})}{c^2}\!+\!\alpha\nu \}
\label{eq:yolo loss function}
\end{equation}
\begin{itemize}
    \item $I\{HD\!\leq\!Th_{HD}\}$: A cut-off function that equals to 1 when condition "$HD\!\leq\!Th_{HD}$" satisfied, and equals to 0 if not.
	\item $IOU$: Ratio of intersection area over union area between two boxes.
	\item $\rho^2 (b,b^{gt})$: Euclidean distance between the centers of two boxes.
	\item $c$: The diagonal of the minimal bounding box that can contain both the predicted box and ground truth.
	\item $\alpha$: trade-off.
	\item $\nu$: Consistency of aspect ratio.
\end{itemize}


\begin{figure}[tb]
	\centering
	\includegraphics[width=80mm]{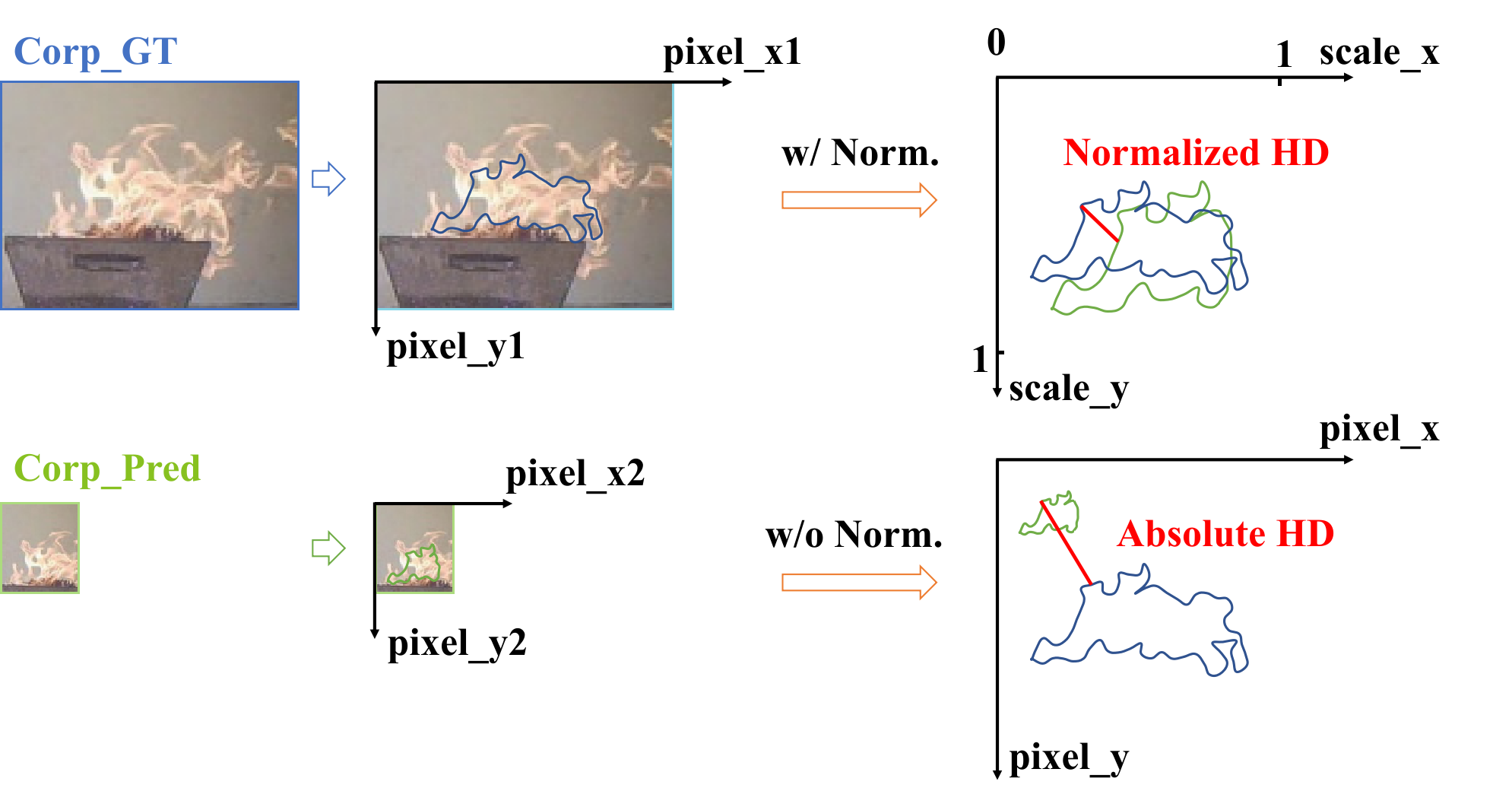}
	\caption{Demonstration of the effect of normalized Hausdorff distance.}
	\label{fig:thre_hd}
\end{figure}

For two-stage Faster R-CNN network, in the first stage, as shown in Figure~\ref{fig:frcn}, it uses a feature pyramid network (FPN) as backbone to get the multi-scale feature maps, and then apply the region proposal network (RPN) to generate a series of object proposals. Afterwards, in the second stage, the region of interest pooling (RoI pooling) layer is used to refine the object proposals towards more accurate bounding boxes. Similarly, we select true false foreground boxes output from the RoI pooling layer to verify their $HD$ with respect to their ground truth, and assign the potential fractions that satisfies $HD\!\leq\!Th_{HD}$ with a lower bounding box loss value (here we set as zero), as described in Equation~\ref{eq:frcn loss function}.

\begin{equation}
L_{bbox}\!=\!I\{HD\!\leq\!Th_{HD}\}\!\cdot\!\sum_{i\in{x,y,w,h}}\!smooth_{L_1}\!(T_i^u\!-\!v_i)
\label{eq:frcn loss function}
\end{equation}
\begin{itemize}
	\item $x,y,w,h$: The central coordinates of the box are denoted by x and y, while its width and height, both measured in pixels, are referred to as w and h, respectively.
	\item $T_i^u$: The predicted values $T$ of the regression parameters $i\in{x,y,w,h}$ for each category $u$.
	\item $v_i$: The ground truth values $v$ of the regression parameters $i\in{x,y,w,h}$.
\end{itemize}

\begin{table*}
	\centering
	\resizebox{\textwidth}{!}{
	\begin{tabular}{l|l|l|lll|lll|lll}
        \toprule
        \multicolumn{2}{c|}{\multirow{2}{*}{\diagbox[height=20pt,innerrightsep=23pt]{Method}{Class}}} & \multirow{2}{*}{Backbone} & \multicolumn{3}{c|}{Division1} & \multicolumn{3}{c|}{Division2} & \multicolumn{3}{c}{Division3} \\
        & & & All & Fire & Smoke & All & Fire & Smoke & All & Fire & Smoke\\
		\midrule
		\multirow{4}{*}{AP50} & YOLOv5\dag & \multirow{2}{*}{CSPDarkNet+SPP} & 0.446 & 0.617 & 0.437 & 0.442 & 0.60 & 0.463 & 0.447 & 0.630 &  0.427\\
		& Ours & & \textbf{0.469} & \textbf{0.624} & \textbf{0.460} & \textbf{0.458} & \textbf{0.627} & \textbf{0.464} & \textbf{0.469} & \textbf{0.648} & \textbf{0.470}\\
        \cmidrule{2-12}
		& Faster R-CNN\dag & \multirow{2}{*}{ResNet-50} & 0.455 & 0.625 & 0.447 & 0.451 & 0.68 & 0.474 & 0.457 & 0.640 &  0.434\\
		& Ours & & \textbf{0.472} & \textbf{0.630} & \textbf{0.463} & \textbf{0.479} & \textbf{0.678} & \textbf{0.481} & \textbf{0.494} & \textbf{0.664} & \textbf{0.470}\\
		\midrule
        \midrule
		\multirow{4}{*}{AP-ss} & YOLOv5\dag & \multirow{2}{*}{CSPDarkNet+SPP} & 0.451 & 0.633 & 0.434 & 0.466 & 0.639 & 0.482 & 0.460 & 0.652 &  0.438\\
		& Ours & & \textbf{0.493} & \textbf{0.661} & \textbf{0.482} & \textbf{0.480} & \textbf{0.654} & \textbf{0.482} & \textbf{0.497} & \textbf{0.671} & \textbf{0.493}\\
		\cmidrule{2-12}
		& Faster R-CNN\dag & \multirow{2}{*}{ResNet-50} & 0.466 & 0.637 & 0.441 & 0.470 & 0.682 & 0.488 & 0.464 & 0.661 & 0.440\\
		& Ours & & \textbf{0.50} & \textbf{0.670} & \textbf{0.489} & \textbf{0.488} & \textbf{0.689} & \textbf{0.495} & \textbf{0.504} & \textbf{0.683} & \textbf{0.497}\\
        \bottomrule
	\end{tabular}
 }
	\caption{Accuracy for three DFS dataset divisions based on standard AP50 and our proposed AP-ss The highest average precision (AP50) scores for each column are displayed in bold, black text. The sign \dag indicates that the method is implemented by us.}
	\label{tab:DFS APs}
\end{table*}

\begin{table}
	\centering
	\resizebox{0.9\linewidth}{!}{
	\begin{tabular}{l|l|l|l}
        \toprule
        \multicolumn{2}{c|}{\diagbox[height=20pt,innerrightsep=23pt]{Method}{Class}} & Backbone & Fire\\
		\midrule
		\multirow{4}{*}{AP50} & YOLOv5\dag & \multirow{2}{*}{CSPDarkNet+SPP} & 0.688\\
		& Ours & & \textbf{0.709}\\
        \cmidrule{2-4}
		& Faster R-CNN\dag & \multirow{2}{*}{ResNet-50} & 0.704\\
		& Ours & & \textbf{0.711}\\
		\midrule
        \midrule
		\multirow{4}{*}{AP-ss} & YOLOv5\dag & \multirow{2}{*}{CSPDarkNet+SPP}& 0.736\\
		& Ours & & \textbf{0.779}\\
		\cmidrule{2-4}
		& Faster R-CNN\dag & \multirow{2}{*}{ResNet-50} & 0.740\\
		& Ours & & \textbf{0.788}\\
        \bottomrule
	\end{tabular}
 }
	\caption{Accuracy for FireNet based on standard AP50 and our proposed AP-ss The highest average precision (AP50) scores for each column are displayed in bold, black text. The sign \dag indicates that the method is implemented by us.}
	\label{tab:FireNet APs}
\end{table}

\subsection{Normalized Hausdorff Distance}

As described in Section~\ref{sec: Self-similar}, $HD$ is used to metric the similarity of two closed contours. The calculation of $HD$ of two images requires to firstly extract the contour from the image.  As shown in Figure ~\ref{fig:thre_hd}, the contour of a fractal (green contour) will exactly match the shape of the contour of the corresponding portion of the whole object (blue contour). Therefore, this approach is not susceptible to any corp operation to the image, which serves as a solid foundation for the development of our proposed methodology.

In order to measure the similarity of two pictures, we extract the maximum connected domain of the image as the represent of this image and as the input contours. Correspondingly, we set a Hausdorff distance threshold ($Th_{HD}$) to determine whether the predicted object have similar shape with target object ($HD\!\leq\!Th_{HD}$). However, such contours are sets of points in the absolute pixel coordinate, which is sensitive to the size of the picture, in other words, absolute $HD$ could not represent the real similarity of two contours if they are not in the same scale. Therefore, we propose to normalize the contours according to the size of the target image so that all points fall in 0 to 1. As shown in Figure ~\ref{fig:thre_hd}, the normalized $HD$ is much smaller than the absolute $HD$ since it could effectively get rid of the impacts of the input image size, and thus better represent the two input similar images. The equation is as described in Equation~\ref{eq:thre_hd}. This strategy also makes it easier to adjust $HD$.

\begin{equation}
\begin{split}
\{X_{norm}^i\}_{max contour} = \{X^i\}_{max contour}/W \\
\{Y_{norm}^i\}_{max contour} = \{Y^i\}_{max contour}/H
\end{split}
\label{eq:thre_hd}
\end{equation}
\begin{itemize}
	\item $X^i, Y^i$: Coordinate of the $i$-th point of the origin contour.
	\item $X_{norm}^i, Y_{norm}^i$: Normalized coordinate of the $i$-th point of the origin contour.
	\item $W, H$: The width and height of the input image.
\end{itemize}

\subsection{Accuracy Milestone}

The timing of using above method during training is also an important factor. If the process of checking self-similar starts too early, then the model do not have the basic ability of recognizing any fire and smoke and consequently may be hard to find the “fractal”; on the contrary, if this process starts too late, then the model may has fall in locally optimal and hard to be tuned. To ensure that the model has a basic level of recognition and is not yet overfitting, we must start our method from an appropriate point in the middle of training, which we call accuracy milestone ($AP_m$). It means when the model reaches certain accuracy insofar as all self-similar objects are concerned ($min(AP50_{fire}, AP50_{smoke}) \geq AP_m$), we start applying our self-similar loss, here we set $AP_m=25\%$. The process of how to get a proper milestone will be discussed in Section~\ref{sec: Experiments}.

\subsection{Self-similar Evaluation Scheme}

Common used object detection accuracy evaluating scheme is AP50, which stands for Average Precision at 50\% IOU (Intersection over Union between the detected object and ground truth). However, according to our problem setting, potential fractals may have lower IOU that smaller than 50\% that should be consider as true positive predictions. Therefore, we propose a new evaluation paradigm to adapt the requirement. Similar to our training strategy, we assess the similarity between the predicted results and the ground truth by calculating the $HD$. If the $HD$ is smaller than the $Th_{HD}$, then this prediction would be positive for calculating the AP, as a result, we nominate this scheme the self-similar AP (AP-ss).

\begin{table}
\begin{center}
	\centering
	\resizebox{0.27\textwidth}{!}{
	\begin{tabular}{l|l|l}
		\toprule
		$A_m$ & $Th_{HD}$ & mAP \\
		\midrule
		5\% & \multirow{6}{*}{Fixed 0.50} & 0.449 \\
		15\% & & 0.445 \\
		25\% & & \textbf{0.469} \\
		35\% & & 0.456 \\
		45\% & & 0.445 \\
		55\% & & 0.443 \\
		\midrule
		\multirow{5}{*}{Fixed 25\%} & 0.05 & 0.453 \\
		& 0.25 & 0.456 \\
		& 0.50 & \textbf{0.469} \\
		& 0.75 & 0.460 \\
		& 0.95 & 0.446 \\
		\bottomrule
	\end{tabular}
     }
\end{center}
\caption{Ablation studies on hyperparameters. The effect of adjusting the accuracy milestone $A_m$ and Hausdorff distance threshold $Th_{HD}$ are demonstrated respectively.}
\label{tab:Ablation}
\end{table}

\begin{figure*}[t]
	\centering
	\includegraphics[width=160mm]{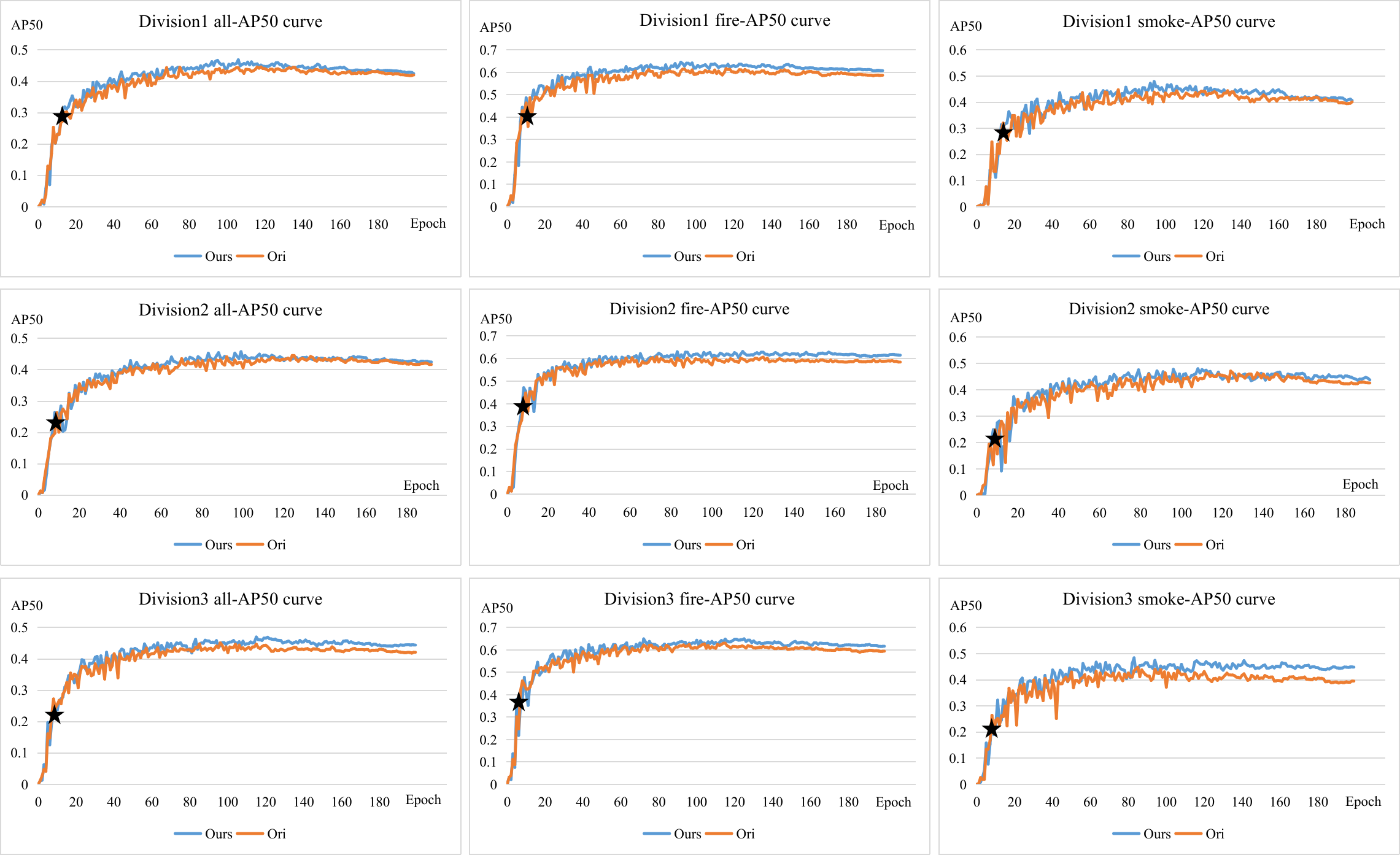}
	\caption{Precision curve over training epochs. The curves in each row refers to DFS division1 to 3. From left to right column are the curve for all categories, fire and smoke separately. The black star indicates our precision milestone.}
	\label{fig:precision curve}
\end{figure*}

\section{Experiments}
\label{sec: Experiments}

As mentioned in Section~\ref{sec:Proposed Methods}, we conduct our experiments on both one-stage and two-stage representative networks. We implement our evaluation on the common used fire and smoke detection dataset: DFS dataset and FireNet dataset. Follow the setting in ~\cite{liu2021real}, we train YOLOv5s and Faster RCNN (ResNet-50 backbone) models from scratch. Additionally, we discuss our hyper-parameter selection and interpretability. Details will be presented as follow.

\subsection{Dataset Partition}

DFS does not provide the public available data set partition. Therefore, we partition the dataset into training, validation and test sets by dividing it into 8:1:1 proportions. As a result, we have 7568 images for training set, 947 images for validation set and 947 images for test set. Additionally, to eliminate the contingency, we repeat this process 3 time with different random seeds to make 3 partitions, namely division1, division2 and division3. The detail of our dataset divisions is presented in the supplementary material and will be made publicly available soon.

\subsection{Results on YOLO}

We conducted the evaluation of our model's performance on the YOLOv5s network using DFS and FireNet. We train the model with single 3090 GPU, the initial learning rate is set as 0.001, using SGD optimizer, total 200 training epochs and batch size is 128 for DFS and 24 for FireNet. 

The evaluation values on DFS is presented in Table~\ref{tab:DFS APs}, which details the results of our model's precision for all categories, fire and smoke on the three dataset divisions. And the evaluation results on FireNet is in Table~\ref{tab:FireNet APs}. We evaluate our model both with the standard average precision (AP50) and our proposed self-similar criteria (AP-ss). As a comparison, we implement the original YOLOv5s network and compare with our method. Our results indicate that our method have greatly improved the detection ability in all cases of all dataset divisions.

\subsection{Results on Faster R-CNN}

Similarly, we perform the same experiment on the Faster R-CNN network. We use four T4 GPUs with batch size of 8, the initial learning rate is 0.001, using SGD optimizer, total 200 training epochs.

The results are reported in Table~\ref{tab:DFS APs} for DFS and Table~\ref{tab:FireNet APs} for FireNet. We also implement the original Faster R-CNN network as comparison. As shown in the table, our model outperform the original network in most cases. We also notice that our reported values on our self-similar criteria show greater improvement.

\subsection{Ablation Study}

In this part we mainly discuss the effect of our hyper parameters, including the accuracy milestone ($A_m$) and the Hausdorff distance threshold ($Th_{HD}$). We use controlled variable and scale $A_m$ from 5\% to 55\%, and scale $Th_{HD}$ from 0.05 to 0.95 to verify the best selection. Our experiment is conducted with our YOLOv5 network and evaluated on DFS division1. The results are as listed in Table~\ref{tab:Ablation}. The highest score occurs when $A_m$=25\% and $Th_{HD}$=0.5, which should be the recommended empirical value.

\begin{figure}[tbp]
	\centering
    \begin{subfigure}{.28\linewidth}
		\includegraphics[height=14mm]{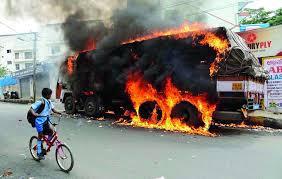}
		\label{fig:large_2990)}
	\end{subfigure}
	\begin{subfigure}{.24\linewidth}
		\includegraphics[height=14mm]{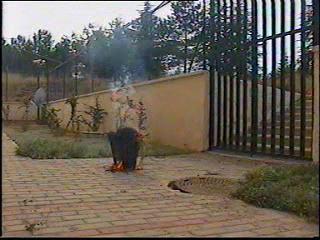}
		\label{fig:middle_2475)}
	\end{subfigure}
	\begin{subfigure}{.26\linewidth}
		\includegraphics[height=14mm]{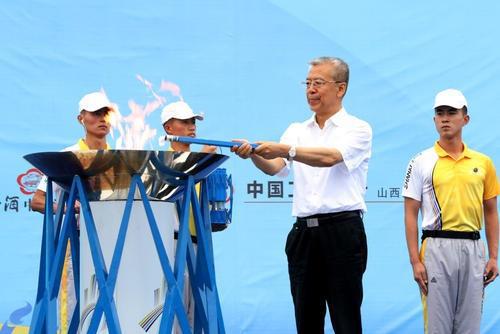}
		\label{fig:middle_488)}
	\end{subfigure}
	\begin{subfigure}{.18\linewidth}
		\includegraphics[height=14mm]{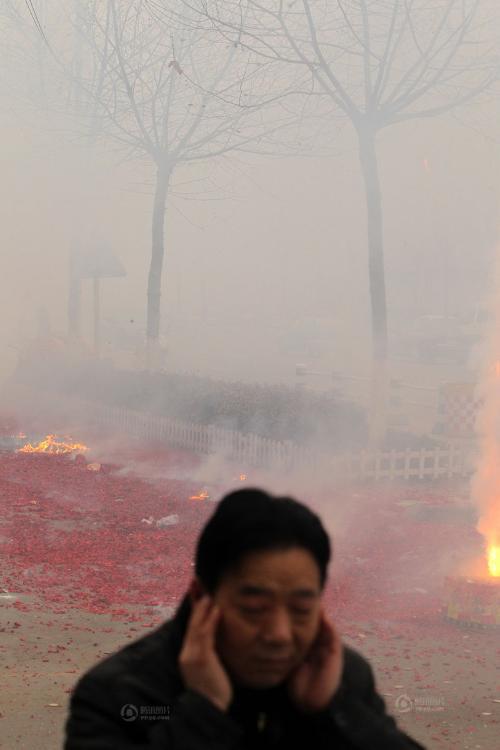}
		\label{fig:middle_2355)}
	\end{subfigure}
    
	\begin{subfigure}{.28\linewidth}
		\includegraphics[height=14mm]{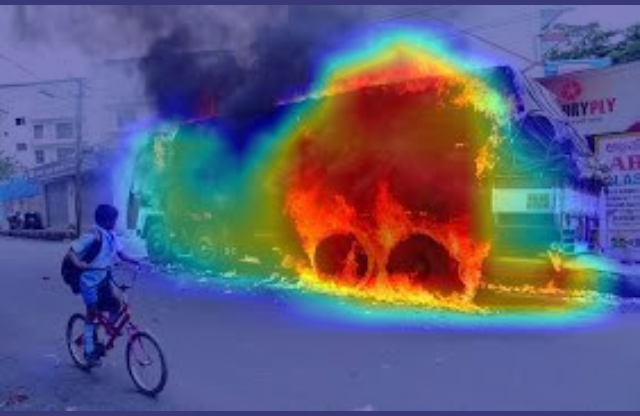}
		\label{fig:gradcam_ori1}
	\end{subfigure}
	\begin{subfigure}{.24\linewidth}
		\includegraphics[height=14mm]{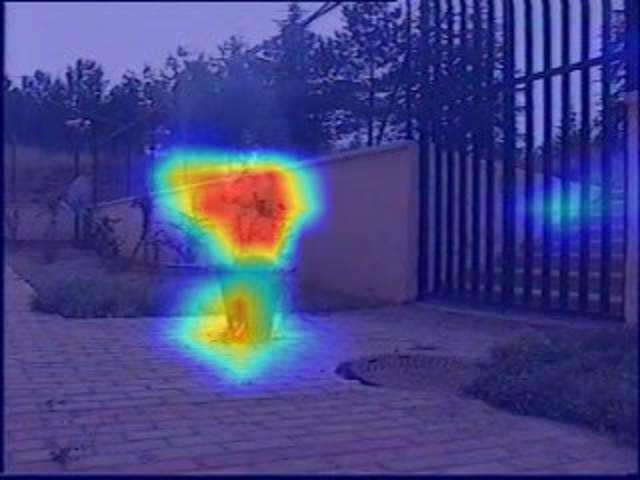}
		\label{fig:gradcam_ori2}
	\end{subfigure}
	\begin{subfigure}{.26\linewidth}
		\includegraphics[height=14mm]{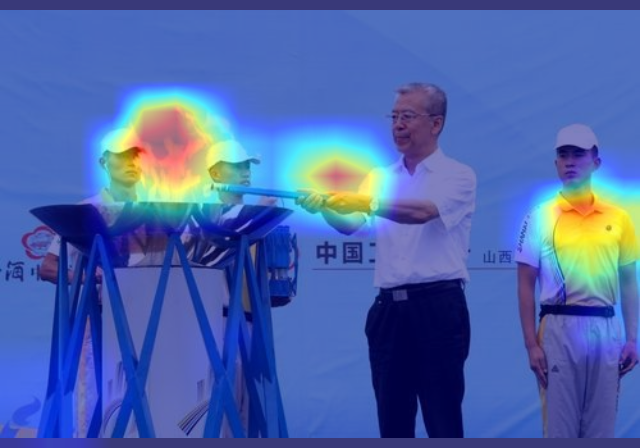}
		\label{fig:gradcam_ori4}
	\end{subfigure}
	\begin{subfigure}{.18\linewidth}
		\includegraphics[height=14mm]{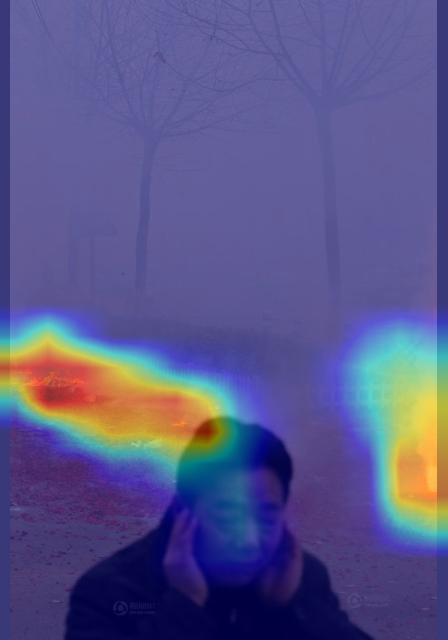}
		\label{fig:gradcam_ori3}
	\end{subfigure}

	\begin{subfigure}{.28\linewidth}
		\includegraphics[height=14mm]{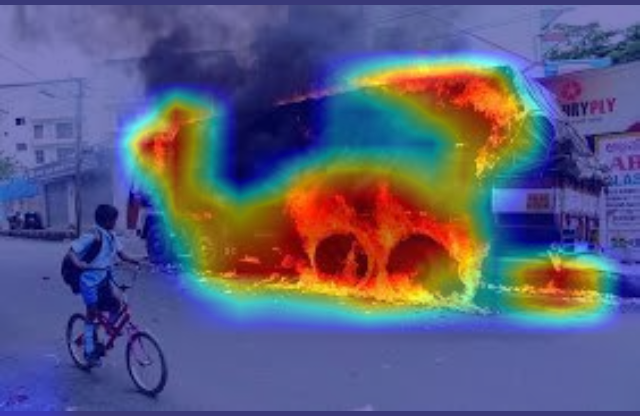}
		\label{fig:gradcam_our1}
	\end{subfigure}
	\begin{subfigure}{.24\linewidth}
		\includegraphics[height=14mm]{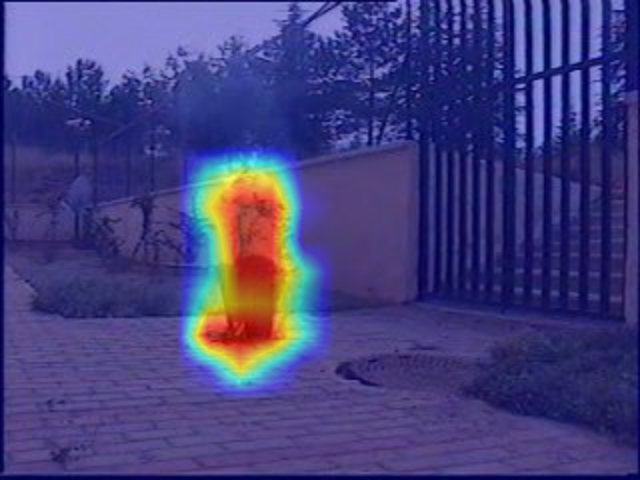}
		\label{fig:gradcam_our2}
	\end{subfigure}
	\begin{subfigure}{.26\linewidth}
		\includegraphics[height=14mm]{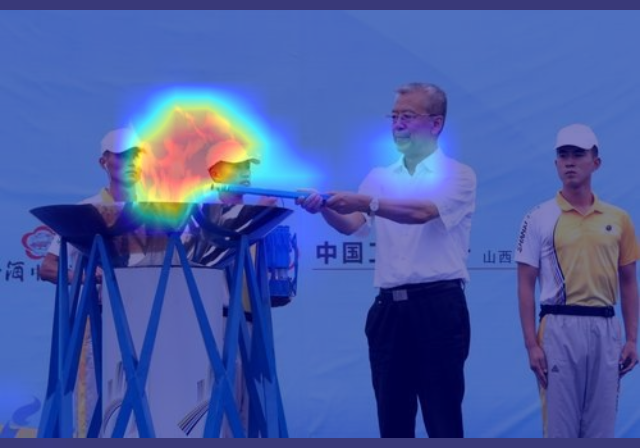}
		\label{fig:gradcam_our4}
	\end{subfigure}
	\begin{subfigure}{.18\linewidth}
		\includegraphics[height=14mm]{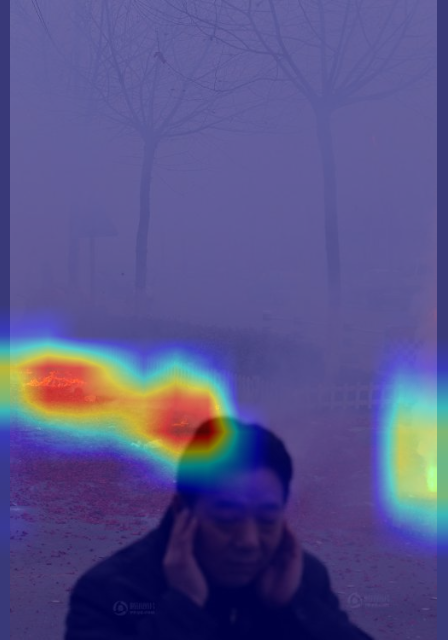}
		\label{fig:gradcam_our3}
	\end{subfigure}
	\caption{Grad-CAM visualization of fire. The color of bright reds and
yellows indicating high importance while cooler blues and greens indicating lower importance. Comparing to the original YOLOv5 network (in the middle row), the performance of our model (in the bottom row) is more precise in identifying the fire and more robust on rejecting the non-fire objects.}
	\label{fig:gradcam}
\end{figure}

\begin{figure}[t]
	\centering
	\begin{subfigure}{0.24\linewidth}
		\centering
		\includegraphics[height=12mm]{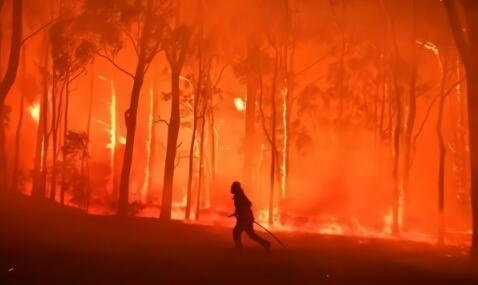}
		\caption{}
		\label{fig:large_411_ori}
	\end{subfigure}
	\begin{subfigure}{0.25\linewidth}
		\centering
		\includegraphics[height=12mm]{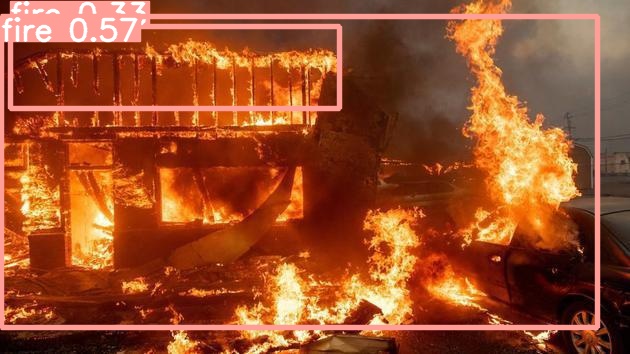}
		\caption{}
		\label{fig:large_2529_ori}
	\end{subfigure}
	\begin{subfigure}{0.24\linewidth}
		\centering
		\includegraphics[height=12mm]{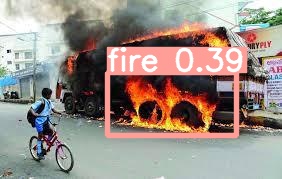}
		\caption{}
		\label{fig:large_2990_ori}
	\end{subfigure}
	\begin{subfigure}{0.2\linewidth}
		\centering
		\includegraphics[height=12mm]{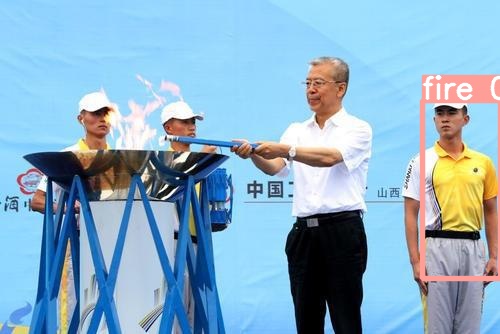}
		\caption{}
		\label{fig:middle_488_ori}
	\end{subfigure}
 
	\begin{subfigure}{0.24\linewidth}
		\centering
		\includegraphics[height=12mm]{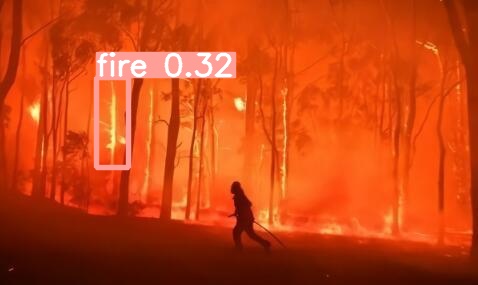}
		\caption{}
		\label{fig:large_411_our}
	\end{subfigure}
	\begin{subfigure}{0.25\linewidth}
		\centering
		\includegraphics[height=12mm]{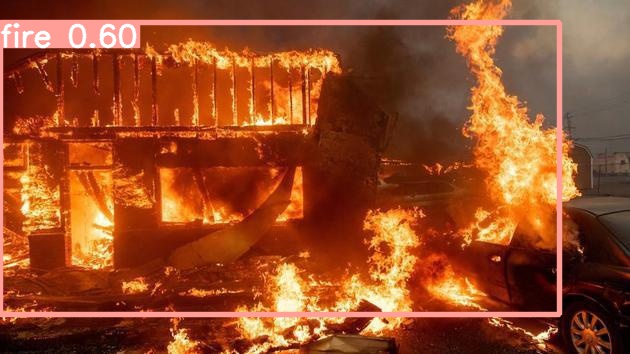}
		\caption{}
		\label{fig:large_2529_our}
	\end{subfigure}
	\begin{subfigure}{0.24\linewidth}
		\centering
		\includegraphics[height=12mm]{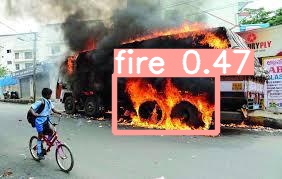}
		\caption{}
		\label{fig:large_2990_our}
	\end{subfigure}
	\begin{subfigure}{0.2\linewidth}
		\centering
		\includegraphics[height=12mm]{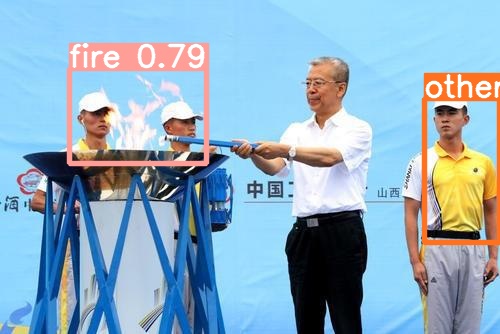}
		\caption{}
		\label{fig:middle_488_our}
	\end{subfigure}
	\caption{Examples of the detection results. Images in the first row demonstrate results of original YOLOv5 model, while the images in the second row are results of our model.}
	\label{fig:detection}
\end{figure}

\begin{table}
\begin{center}
	\centering
	\resizebox{0.3\textwidth}{!}{
	\begin{tabular}{l|l|l}
		\toprule
		Method & \makecell{Inference\\speed (ms)} & \makecell{Trainable\\parameters} \\
		\midrule
		YOLOv5\dag & 6.2 & 7,235,389 \\
		Ours & 5.8 & 7,235,389 \\
		\bottomrule
	\end{tabular}
 }
\end{center}
\caption{The computation overhead of our model. The sign \dag indicates that the method is implemented by us.}
\label{tab:inference}
\end{table}

\subsection{Other Experiments}

In this part we discuss the interpretability and other model performance, all experiments below are based on DFS division1 test set, using YOLOv5s network if not specified.

\textbf{Interpretability}
To further illustrate the effect of our method, we implement the Grad-CAM visualization on the original YOLOv5 network and our improved method. Grad-CAM uses the gradients of target feature maps to obtain feature map weights, which can be viewed as the contribution of each pixel in the gradient map to the final decision~\cite{8237336,8354201}. We present several images and their Grad-CAM visualization for the fire category, as seen in Figure~\ref{fig:gradcam}: the original YOLOv5 model is difficult to notice the precise fire area and even makes wrong attention, while our improved model could correctly and precisely identify the fire area.

\textbf{Detection Results}
In order to intuitively demonstrate the improvement of our model, some representative inference results are as list in Figure~\ref{fig:detection}. From left column to right, we present our improvement on missing predictions, not unified boxes, low confidence and wrong predictions.


\textbf{Precision Curve}
We present the precision curve in Figure~\ref{fig:precision curve} based on three DFS dataset divisions. We mark the accuracy milestone on the curve, which is the time our self-similar loss is applied; as indicate in the figure, our method is able to improve the detection precision in all cases.


\textbf{Inference Speed}
We also reported the inference speed and model size of our model. As shown in Table~\ref{tab:inference} we do not add extra weights or convolutional modules, our model size is as same as the original YOLOv5 model.

\begin{table}
\begin{center}
	\centering
	\resizebox{0.3\textwidth}{!}{
	\begin{tabular}{l|l|l|l}
		\toprule
		Method & All & Fire & Smoke \\
		\midrule
		YOLOv5\dag & 0.462 & 0.596 & 0.445\\
		Ours & \textbf{0.485} & \textbf{0.650} & \textbf{0.468}\\
		\bottomrule
	\end{tabular}
    }
\end{center}
\caption{The recall value of our model for each category. The sign \dag indicates that the method is implemented by us.}
\label{tab:recall}
\end{table}

\textbf{Recall} Additionally, to evaluate the ability of our proposed method in finding out the true fractals, we conduct the evaluation of recall of our model, the result is  shown in Table~\ref{tab:recall}. Our model have higher recall value, which indicates our method could propose more positive predictions.

\subsection{Discussion}

Our experiments indicates that self-similar, as an important feature of fire and smoke, could be well-explored in improving the detection ability. Hausdorff distance, as one of the intuitive methods in determining the similarity between fractals and whole, is heavily relies on high-quality contours that extract from the images, and therefore could be further improved by more precise methods of getting contours. Through our proposed semi-supervised learning scheme, more fractals of fire and smoke are proposed, which effectively prevents the model from making incorrect predictions on potential positive samples.

In addition, to be expected, our method is not only valid in fire and smoke detection, but also can be easily adapt to other tasks that have self-similar pattern such as industrial defect detection, dust detection and coastline detection. 


\section{Conclusion}

In this paper, we first propose a novel semi-supervised network to address the object detection task that with distinctive self-similar feature. Our proposed self-similar loss based on Hausdorff distance also demonstrates strong ability in identifying the potential fractals of fire and smoke. Additionally, as a specific evaluating scheme, we report the result on our proposed AP-ss as a new comparable benchmark for future works. Our experimental results indicates that our methods is simple yet effective, and have surpassed the existing baseline methods. We hope our method could bring benefit on more potential fire disaster control project.

\bibliography{aaai24}

\end{document}